%% file: SelfMGNN.tex
\def\year{2022}\relax
\documentclass[letterpaper]{article} % DO NOT CHANGE THIS
\usepackage{aaai22}  % DO NOT CHANGE THIS
\usepackage{times}  % DO NOT CHANGE THIS
\usepackage{helvet}  % DO NOT CHANGE THIS
\usepackage{courier}  % DO NOT CHANGE THIS
\usepackage[hyphens]{url}  % DO NOT CHANGE THIS
\usepackage{graphicx} % DO NOT CHANGE THIS
\usepackage{natbib}  % DO NOT CHANGE THIS AND DO NOT ADD ANY OPTIONS TO IT
\usepackage{caption} % DO NOT CHANGE THIS AND DO NOT ADD ANY OPTIONS TO IT
\DeclareCaptionStyle{ruled}{labelfont=normalfont,labelsep=colon,strut=off} % DO NOT CHANGE THIS
\usepackage[ruled,vlined]{algorithm2e}

\usepackage{algorithmic}

\usepackage{amsmath}
\usepackage{bm}
\usepackage{mathrsfs}
\usepackage{amsthm}
\usepackage{amsfonts}
\usepackage{subfigure}
\usepackage{multirow}
\usepackage{url}
\usepackage{array}
\usepackage{amssymb}
\usepackage{booktabs} 

\usepackage{color}
\newcommand{\blue}[1]{{{\textcolor{blue}{#1}}}}
\title{A Self-supervised Mixed-curvature Graph Neural Network}
\author{
Li Sun\textsuperscript{\rm 1}\thanks{Corresponding Author: Li Sun, ccesunli@ncepu.edu.cn},  Zhongbao Zhang\textsuperscript{\rm 2}, Junda Ye\textsuperscript{\rm 2}, Hao Peng\textsuperscript{\rm 3}, Jiawei Zhang\textsuperscript{\rm 4}, \\Sen Su\textsuperscript{\rm 2}  and Philip S. Yu\textsuperscript{\rm 5}. 
}
\affiliations{
    \textsuperscript{\rm 1}School of Control and Computer Engineering, North China Electric Power University, Beijing 102206, China\\
    \textsuperscript{\rm 2}School of Computer Science, Beijing University of Posts and Telecommunications, China\\
    \textsuperscript{\rm 3}Beijing Advanced Innovation Center for Big Data and Brain Computing, Beihang University, Beijing 100191, China\\
    \textsuperscript{\rm 4}IFM Lab, Department of Computer Science, University of California, Davis, CA, USA\\
    \textsuperscript{\rm 5}Department of Computer Science, University of Illinois at Chicago, IL, USA\\
    ccesunli@ncepu.edu.cn; \{zhongbaozb, susen\}@bupt.edu.cn; penghao@act.buaa.edu.cn; \\jiawei@ifmlab.org; psyu@uic.edu

}
\usepackage{bibentry}
\begin{document}

\maketitle

\input{abstract}

\input{intro}

\input{preliminary}

\input{methodology}

\input{experiments}

\input{relatedwork}

\section{Acknowledgments}
Hao Peng is partially supported by the National Key R\&D Program of China through grant 2021YFB1714800,  NSFC through grant 62002007, S\&T Program of Hebei through grant 21340301D.
Sen Su and Zhongbao Zhang are partially supported by the National Key Research and Development Program of China under Grant 2018YFB1003804, NSFC under Grant U1936103 and 61921003.
Philip S. Yu is partially supported by NSF under grants III-1763325, III-1909323,  III-2106758, and SaTC-1930941. 
Jiawei Zhang is partially supported by NSF through grants IIS-1763365, IIS-2106972 and by UC Davis.
This work was also sponsored by CAAI-Huawei MindSpore Open Fund.
Thanks for computing infrastructure provided by Huawei MindSpore platform.

\section{Technical Appendix}
\input{appendix}

% Use \b{}ibliography{yourbibfile} instead or the References section will not appear in your paper
\bibliography{aaai22}

% \section{Acknowledgments}
% AAAI is especially grateful to Peter Patel Schneider for his work in implementing the original aaai.sty file, liberally using the ideas of other style hackers, including Barbara Beeton. We also acknowledge with thanks the work of George Ferguson for his guide to using the style and BibTeX files --- which has been incorporated into this document --- and Hans Guesgen, who provided several timely modifications, as well as the many others who have, from time to time, sent in suggestions on improvements to the AAAI style. We are especially grateful to Francisco Cruz, Marc Pujol-Gonzalez, and Mico Loretan for the improvements to the Bib\TeX{} and \LaTeX{} files made in 2020.

% The preparation of the \LaTeX{} and Bib\TeX{} files that implement these instructions was supported by Schlumberger Palo Alto Research, AT\&T Bell Laboratories, Morgan Kaufmann Publishers, The Live Oak Press, LLC, and AAAI Press. Bibliography style changes were added by Sunil Issar. \verb+\+pubnote was added by J. Scott Penberthy. George Ferguson added support for printing the AAAI copyright slug. Additional changes to aaai22.sty and aaai22.bst have been made by Francisco Cruz, Marc Pujol-Gonzalez, and Mico Loretan.

% \bigskip
% \noindent Thank you for reading these instructions carefully. We look forward to receiving your electronic files!

\end{document}

%% file: abstract.tex
%!TEX root = ./SelfMGNN.tex

\begin{abstract}
Graph representation learning received increasing attentions in recent years.
Most of the existing methods ignore the complexity of the graph structures and restrict graphs in a single constant-curvature representation space, which is only suitable to particular kinds of graph structure indeed.
Additionally, these methods follow the supervised or semi-supervised learning paradigm, and thereby notably limit their deployment on the unlabeled graphs in real applications.
To address these aforementioned limitations, 
we take the first attempt to study the self-supervised graph representation learning in the mixed-curvature spaces.
In this paper, 
we present a novel \textbf{Self}-supervised \textbf{M}ixed-curvature \textbf{G}raph \textbf{N}eural \textbf{N}etwork (\textbf{\textsc{SelfMGNN}}).
To capture the complex graph structures, 
we construct a \emph{mixed-curvature}  space via the Cartesian product of multiple Riemannian component spaces, 
and design hierarchical attention mechanisms for learning and fusing graph representations across these component spaces. 
% To enable the self-supervisd learning, we propose a novel \emph{dual contrastive approach} that con.
% Specifically, we first introduce a Riemannian projector to reveal different Riemannian views. 
% Then, we utilize a well-designed Riemannian discriminator for 
% single-view contrastive learning, contrasting positive and negative samples in the same Riemannian view, 
% and cross-view contrastive learning, contrasting between different Riemannian views concurrently.
To enable the self-supervisd learning, we propose a novel \emph{dual contrastive approach}.
The constructed mixed-curvature space actually provides multiple Riemannian views for the contrastive learning. 
We introduce a Riemannian projector to reveal these views,
 and utilize a well-designed Riemannian discriminator for the \emph{single-view} and \emph{cross-view contrastive learning} within and across the Riemannian views.
Finally, extensive experiments show that \textsc{SelfMGNN} captures the complex graph structures and outperforms state-of-the-art baselines.
\end{abstract}

%% file: intro.tex
%!TEX root = ./SelfMGNN.tex

\section{Introduction}

Graph representation learning \cite{cui2018survey,hamilton2017inductive} shows fundamental importance in various applications, 
such as link prediction and node classification \cite{kipf2016semi}, 
and thus receives increasing attentions from both academics and industries.
%We discuss the limitations of prio,r works in the following aspects:
Meanwhile, we have also observed great limitations with the existing graph representation learning methods in two major perspectives, 
which are described as follows:

%Uni-curvature:
\noindent \textbf{\emph{Representation Space}}:
Most of existing methods ignore the complexity of real graph structures, and limit the graphs in a single \emph{constant-curvature} representation space \cite{GuSGR19}.
Such methods can only work well on particular kinds of structure that they are designed for.
% The (zero-curvature) Euclidean space has been the workhorse for graph representation learning \cite{}.
% Recently, the Riemannian space, such as hyperbolic or spherical space, 
% has gained increasing attentions by providing better representations for certain types of data, 
For instance, the constant negative curvature hyperbolic space is well-suited for graphs with hierarchical or tree-like structures \cite{HGNN}.
The constant positive curvature spherical space is especially suitable for data with cyclical structures, e.g., triangles and cliques \cite{BachmannBG20},
and the zero-curvature Euclidean space for grid data \cite{WuPCLZY21}.
% For better representing general-structured graphs, it calls for a \emph{mixed curvature} representation space in fact.
% The motivation is intuitive:
However, graph structures in reality are usually mixed and complicated rather than uniformed, in some regions hierarchical, while in others cyclical \cite{papadopoulos2012popularity,ravasz2003hierarchical}.
Even more challenging, the curvatures over different hierarchical or cyclical regions can be different as will be shown in this paper.
%Thus, it calls for  a \emph{mixed-curvature} space to match the wide variety of graph structures for providing more promising representations.
In fact, it calls for a new representation space to match the wide variety of graph structures, and we seek spaces of \emph{mixed-curvature} to provide better representations.
% Self-supervised:
% Self-supervised graph representation learning is a more favorable choice in many cases due to the freedom from labels, particularly when we intend to take advantage from a large scale unlabeled graph in the wild.

%training GNNs in existing approaches usually requires a certain form of supervision.
\noindent \textbf{\emph{Learning Paradigm}}:
Learning graph representations usually requires abundant supervision label information \cite{velickovic2018graph,HGCN}.
Labels are usually scarce in real applications, 
and undoubtedly, labeling graphs is expensive—manual annotation or paying for permission, and is even impossible to acquire because of the privacy policy.
Fortunately, the rich information in graphs provides the potential for \emph{self-supervised learning}, i.e., learning representations without labels \cite{DBLP:journals/corr/abs-2006-08218}. 
Self-supervised graph representation learning is a more favorable choice, 
particularly when we intend to take the advantages from the unlabeled graphs in real applications.
% Recently, a few attempts \cite{VelickovicFHLBH19,HassaniA20,QiuCDZYDWT20} show that \emph{contrastive learning}, i.e., learning graph representations by contrasting graphs with the congruent or discongruent counterpart, is a promising method.
% %However, to the best of knowledge, none of existing Riemannian GNNs equip the ability of self-supervised learning. 
Recently, contrastive learning \cite{VelickovicFHLBH19,QiuCDZYDWT20}
emerges as a successful method for the graph self-supervised learning.
However, existing self-supervised methods, 
to the best of our knowledge, 
cannot be applied to the mixed-curvature spaces due to the intrinsic differences in the geometry.

To address these aforementioned limitations, 
we take the first attempt to study the \emph{self-supervised graph representation learning in the mixed-curvature space} in this paper. 

To this end, we present a novel \textbf{Self}-supervised \textbf{M}ixed-curvature \textbf{G}raph \textbf{N}eural \textbf{N}etwork, named \textbf{\textsc{SelfMGNN}}. %referred to as \textsc{SelfMGNN}.
%To address the first limitation, we model the graphs in a representation space of mixed curvature.
To address the first limitation, 
we propose to learn the representations in a \emph{mixed-curvature space}.
Concretely, we first construct a mixed-curvature space via the Cartesian product of multiple Riemannian—hyperbolic, spherical and Euclidean—component spaces, jointly enjoying the strength of different curvatures to match the complicated graph structures.
%The mixed-curvature space has multiple hyperbolic (spherical) components with learnable curvatures.  
%We utilize the $\kappa$-stereographic model to give the unified formalism of an arbitrary hyperbolic or spherical component spaces. %, i.e., a positive $\kappa$ for spherical, a negative $\kappa$ for hyperbolic, and the limit at zero for Euclidean.
Then, we introduce hierarchical attention mechanisms for learning and fusing representations in the product space.
In particular, we design an intra-component attention for the learning within a component space
and an inter-component attention for the fusing across component spaces.
To address the second limitation, 
we propose a novel \emph{dual contrastive approach} to enable the self-supervisd learning.
The constructed mixed-curvature space %not only matches the complicate structures of graphs, 
actually provides multiple Riemannian views for contrastive learning.
Concretely, we first introduce a Riemannian projector to reveal these views, i.e., hyperbolic, spherical and Euclidean views. 
Then, we introduce the \emph{single-view} and \emph{cross-view contrastive learning}. % are performed simultaneously.
In particular, we utilize a well-designed Riemannian discriminator to 
contrast positive and negative samples in the same Riemannian view  (i.e., the single-view contrastive learning)
and concurrently contrast between different Riemannian views (i.e., the cross-view contrastive learning).
In the experiments, we study the curvatures of real graphs and show the advantages of allowing multiple positive and negative curvature components for the first time,  
%graph in the mixed-curvature spacedecomposition of the Cartesian product space for self-supervised GNNs for the first time and 
demonstrating the superiority of \textsc{SelfMGNN}.
% to capture the complicated graph structures without labels.

Overall, our main contributions are summarized below:
\begin{itemize}
\item \emph{Problem}: To the best of our knowledge, this is the first attempt to study the self-supervised graph representation learning in the mixed-curvature space.
\item \emph{Model}: This paper presents a novel \textsc{SelfMGNN} model, where hierarchical attention mechanisms and dual contrastive approach are designed for self-supervised learning in the mixed-curvature space, allowing multiple hyperbolic (spherical) components with distinct curvatures.  
%\item \emph{Experiments}: Extensive experiments show the superiority of \textsc{SelfMGNN}. This paper is the first to show the advantage of allowing multiple positive (negative) curvature components for graph in the mixed-curvature space, demonstrating that the curvatures over the different hierarchical (spherical) regions of a graph can be different.  
\item \emph{Experiments}: Extensive experiments show the curvatures over different hierarchical (spherical) regions of a graph can be different.  \textsc{SelfMGNN} captures the complicated graph structures without labels and outperforms the state-of-the-art baselines. 

\end{itemize}

%% file: preliminary.tex
%!TEX root = ./SelfMGNN.tex

\section{Preliminaries and Problem Definition}

% Prior to introducing the studied problem and proposed approach, we provide some important preliminaries in this section. 
% Throughout the paper, 
% we denote the Euclidean norm and inner product by $\| \cdot \|_2$ and $\langle \cdot, \cdot \rangle$, respectively.
In this section, we first present the preliminaries and notations necessary to construct a mixed-curvature space.
%(more details can be found in textbooks [28, 39]), 
%including the Riemannian manifold and constant-curvature space. 
Then, we formulate the problem of \emph{self-supervised graph representation learning in the mixed-curvature space}.

\subsection{Riemannian Manifold}

A smooth \emph{manifold} $\mathcal M$ generalizes the notion of the surface to higher dimensions.
Each point $\mathbf x \in \mathcal M$ associates with a \emph{tangent space} $\mathcal T_\mathbf x\mathcal M$, the first order approximation of $\mathcal M$ around $\mathbf x$, which is locally Euclidean.
On tangent space $\mathcal T_\mathbf x\mathcal M$, the \emph{Riemannian metric}, $g_\mathbf x (\cdot, \cdot) : \mathcal T_\mathbf x\mathcal M  \times \mathcal T_\mathbf x\mathcal M \to \mathbb R$, defines an inner product so that geometric notions can be induced.
The tuple $(\mathcal M, g)$ is called a \emph{Riemannian manifold}.

Transforming between the tangent space and the manifold is done via exponential and logarithmic maps, respectively.
For $\mathbf x \in \mathcal M$, 
the  \emph{exponential map} at $\mathbf x$, 
$\mathbf{exp}_\mathbf x(\mathbf v): \mathcal T_\mathbf x\mathcal M \to \mathcal M$, 
projects the vector $\mathbf v \in \mathcal T_\mathbf x\mathcal M$ onto the manifold $\mathcal M$.
The \emph{logarithmic map} at $\mathbf x$, 
$\mathbf{log}_\mathbf x(\mathbf y): \mathcal M \to \mathcal T_\mathbf x\mathcal M$, 
projects the vector $\mathbf y \in \mathcal M$ back to the tangent space $\mathcal T_\mathbf x\mathcal M$.
For further expositions, please refer to mathematical materials \cite{Spivak1979,Hopper2010}.

\subsection{Constant Curvature Space}

The Riemannian metric also defines a curvature at each point $\kappa(\mathbf x)$, 
which determines how the space is curved.
If the curvature is uniformly distributed,  
$(\mathcal M, g)$ is called a \emph{constant curvature space} of curvature $\kappa$. 
There are $3$ canonical types of constant curvature space that we can define with respect to the sign of the curvature: 
a positively curved spherical space $\mathbb S$ with $\kappa>0$, 
a negatively curved hyperbolic space $\mathbb H$ with $\kappa<0$ 
and the flat Euclidean space $\mathbb E$  with $\kappa=0$.

\noindent\textbf{Note that}, $\| \cdot \|_2$ denotes the Euclidean norm in this paper.

\subsection{Problem Definition}
In this paper, we propose to study the self-supervised graph representation learning in the mixed-curvature space.
Without loss of generality,
a graph is described as $G = (V, E, \mathbf X)$, 
where $V = \{v_1,  \cdots, v_n\}$ is the node set and $E =\{ (v_i,  v_j ) | \  v_i,  v_j \in V\}$ is the edge set. 
We summarize the edges in the adjacency matrix $\mathbf G$, where $\mathbf G_{ij}=1$ iff $(v_i,  v_j ) \in E$, otherwise $0$.
Each node $v_i$ is associated with a feature vector $\mathbf x_i \in \mathbb R^d$, and matrix $\mathbf{X} \in \mathbb{R}^{|V| \times d}$ represents the features of all nodes.
Now, we give the studied problem:
\newtheorem*{def1}{Problem Definition (Self-supervised graph representation learning in the mixed-curvature space)} 
\begin{def1}
Given a graph $G = (V, E, \mathbf X)$, 
the problem of self-supervised graph representation learning in the mixed-curvature space
is to learn an encoding function $\Phi: V \to \mathcal P$ that maps the node $v$ to a vector $\boldsymbol z$ in a mixed-curvature space $\mathcal P$ that captures the intrinsic complexity of graph structure without using any label information. 
\end{def1}

In other words, the graph representation model should align with the complex graph structures 
— hierarchical as well as cyclical structure, 
and can be learned without external guidance (labels). 
% The representation space of constant curvature works well on particular kinds of structure that they were designed for. % \cite{HGNN,ZhangWSLS21}.
% Instead of the constant-curvature space, the mixed-curvature space is called in fact to cover the intrinsic complicated structures of the graphs.
Graphs in reality are usually mixed-curvatured rather than structured uniformly, i.e., in some regions hierarchical, while in others cyclical.
A constant-curvature model (e.g., hyperbolic, spherical or the Euclidean model) benefits from their specific bias to better fit particular structure types. 
To bridge this gap, we propose to work with the \textbf{mixed-curvature space} to cover the  complex graph structures in real-world applications.

%% file: methodology.tex
%!TEX root = ./SelfMGNN.tex

\begin{figure*}
\centering
    \includegraphics[width=1\linewidth]{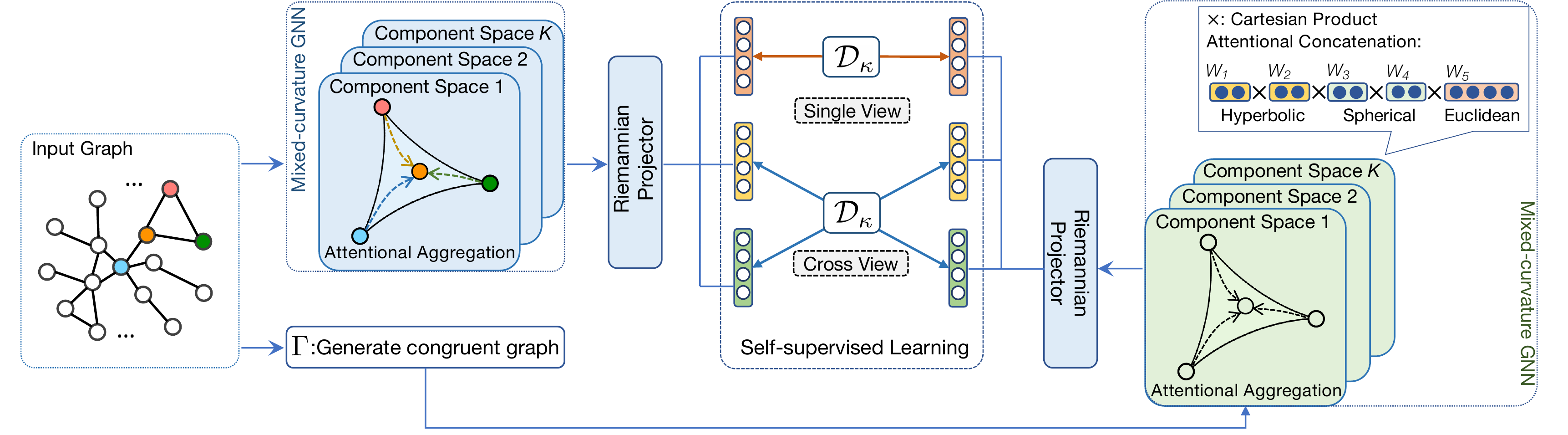}
    \caption{Overall architecture of \textsc{SelfMGNN}: \footnotesize{In \textsc{SelfMGNN}, we first introduce a mixed-curvature GNN to learn graph representations. 
    Specifically, we perform attentional aggregation within the component space where the triangle is to show its geometry, e.g., triangles curve inside in $\mathbb H$,  and attentional concatenation among component spaces whose example with learnable weights is on the top right. 
    Then, to enable self-supervised learning, we design a Riemannian projector to reveal different views of the mixed-curvature space, and utilize a well-designed Riemannian discriminator $\mathcal D_\kappa$ to contrast samples for single- and cross-view contrastive learning, shown in the middle. 
    In practice, we feed the graph and its congruent augmentation, generated by $\Gamma(\cdot)$, for the contrastive learning as specified in Algo. \ref{algo}. }}
    \label{illu}
\end{figure*}

%\section{\textsc{SelfMGNN:} Self-supervised Mixed-curvature GNN}

\section{\textsc{SelfMGNN}: Our Proposed Model}

To address this problem, we present a novel \textbf{Self}-supervised \textbf{M}ixed-curvature \textbf{G}raph \textbf{N}eural \textbf{N}etwork (\textbf{\textsc{SelfMGNN}}).
In a nutshell, 
\textsc{SelfMGNN} learns graph representations in the mixed-curvature space, 
and is equipped with a dual contrastive approach to enable its self-supervised learning. 
We illustrate the architecture of \textsc{SelfMGNN} in Fig. \ref{illu}.
We will elaborate on the mixed-curvature graph representation learning and the dual contrastive approach in following sections.
% $\mathbf \Phi: v \to \mathbf z \in \mathcal P^d$
% \noindent  We introduce a hierarchical attention mechanism to update node representations in the product space.

\subsection{Mixed-curvature GNN}

%a hierarchical attention mechanism in the mixed-curvature space
%The mixed-curvature GNN is a generic graph neural network 
To learn the graph representations in a mixed-curvature space,
we construct a mixed-curvature space by the Cartesian product of multiple Riemannian component spaces,
in which we propose a mixed-curvature GNN with hierarchical attention mechanisms.
In particular, we first stack \emph{intra-component attentional layers} in each component space to learn constant-curvature component representations.
Then, we design an \emph{inter-component attentional layer} across component spaces to fuse these component representations
%, so that the output encodings better match the intrinsic heterogeneous structures of the graphs.
 so as to obtain the output mixed-curvature representations matching the complex graph structures.

\subsubsection{Constructing a Mixed-Curvature Space:}
We leverage the \emph{Cartesian product} to construct the mixed-curvature space.
With $K$ constant-curvature spaces $\{\mathcal M_i\}_{i=1}^K$ indexed by subscript $i$, 
we perform the Cartesian product over them and obtain the resulting product space $\mathcal P = \times_{i=1}^K \mathcal M_i$,
where $\times$ denotes the Cartesian product, and $\mathcal M_i$ is  known as a component space.
By fusing multiple constant-curvature spaces, 
the product space is constructed with non-constant mixed curvatures, matching the complex graph structures.  

The product space $\mathcal P$ with the dimensionality $d$ is described by its \emph{signature}, which has three degrees of freedom per component: 
i) the model $\mathcal M_i$, 
ii) the dimensionality $d_i$  and 
iii) the curvature $\kappa_i$, where $\sum\nolimits_{i=1}^K d_i=d$.
We use a shorthand notation for repeated components, $(\mathcal M_k)^j=\times_{i=1}^j \mathcal M_k$.
Note that, \textsc{SelfMGNN} can have multiple hyperbolic or spherical components with distinct learnable curvatures, and such a design enables us to cover a wider range of curvatures for the better representation.
However, we only need one Euclidean space, since the Cartesian product of Euclidean space is $\mathbb E^{d_0}=\times_{i=1}^j \mathbb E_i$, 
and $\sum\nolimits_{i=1}^j d_i=d_0$.
%The equality does not hold for hyperbolic or spherical spaces. 

The Cartesian product introduces a simple and interpretable combinatorial construction of the mixed-curvature space.
For $\mathbf x=||_{i=1}^K \mathbf x_{\mathcal M_i}$ in product space $\mathcal P$, $\mathbf x_{\mathcal M_i}$ denotes the component embedding in  $\mathcal M_i$ and $||$ denotes the vector concatenation.
Thanks to the combinatorial construction, we can first learn representations in each component space and then fuse these representations in the product space.

\subsubsection{Model and Operations:}
Prior to discussing about the learning and fusing of the representations, we give the model of component spaces and provide the generalized Riemannian operations in the component spaces in this part.

We opt for the \emph{$\kappa$-stereographic model} as 
it  
\emph{unifies spaces of both positive and negative curvature} and 
\emph{unifies operations with gyrovector formalism.}
Specifically,  the $\kappa$-stereographic model is a smooth manifold $\mathcal M^d_\kappa=\{\boldsymbol z \in \mathbb R^d  | -\kappa||\boldsymbol z ||_2^2 < 1\}$, whose origin is $\mathbf 0 \in \mathbb R^d$,
equipped with a Riemannian metric 
$g_{\boldsymbol z}^\kappa=(\lambda_{\boldsymbol z}^\kappa)^2 \mathbf I$, 
where $\lambda_{\boldsymbol z}^\kappa$ is given below:
\begin{equation}
\lambda_{\boldsymbol z}^\kappa=2\left(1+\kappa||\boldsymbol z||_2^2\right)^{-1}.
\label{conformalfactor}
\end{equation}
%The $\kappa$-stereographic model is \emph{a unified model regardless of the sign of constant curvature $\kappa$}. 
In particular,  $\mathcal M^d_\kappa$ is the stereographic sphere model for spherical space ($\kappa > 0$),
while it is the Poincar\'e ball model of radius $1/ \sqrt{-\kappa}$ for hyperbolic space ($\kappa < 0$).
% The $\kappa$-stereographic model generalizes gyrovector spaces to positive curvature, 
% so that we have \emph{unified operation with gyrovector formalism}, 
% i.e., an elegant non-associative algebra in analogy with the Euclidean space.
We summarize all the necessary operations for this paper in Table \ref{tab:ops} 
with the curvature-aware definition of trigonometric functions. 
Specifically, $\tan_\kappa(\cdot)=\tan(\cdot)$ if $\kappa<0$ and $\tan_\kappa(\cdot)=\tanh(\cdot)$ if $\kappa>0$.
Note that, the bold letter denotes the vector on the manifold.

\begin{table*}
  %\scriptsize
\centering
\caption{Summary of the operations in constant-curvature space (hyperbolic $\mathbb H^d$, spherical $\mathbb S^d$ and Euclidean space $\mathbb E^d$).}
\begin{tabular}{l|c|c}
\hline
\textbf{Operation} & \textbf{Formalism in $\mathbb E^d$} &\textbf{Unified formalism in $\kappa$-stereographic model ($\mathbb H^d$/ $\mathbb S^d$)}\\
\hline
Distance Metric& 
$d^\kappa_{\mathcal M}(\mathbf{x}, \mathbf{y}) =\left\| \mathbf{x}- \mathbf{y}\right\|_{2}$
&
$
d^\kappa_{\mathcal M}(\mathbf{x}, \mathbf{y})=\frac{2}{\sqrt{|\kappa|}} \tan _{\kappa}^{-1}\left(\sqrt{|\kappa|}\left\|-\mathbf{x} \oplus_{\kappa} \mathbf{y}\right\|_{2}\right)
$
\\
\hline
Exponential Mapping & 
$\mathbf{exp}_{\mathbf{x}}^{\kappa}(\mathbf{v})=\mathbf{x}+\mathbf{v}$ 
&
$
\mathbf{exp}_{\mathbf{x}}^{\kappa}(\mathbf{v})=\mathbf{x} \oplus_{\kappa}\left(\tan _{\kappa}\left(\sqrt{|\kappa|} \frac{\lambda_{\mathbf{x}}^{\kappa}\|\mathbf{v}\|_{2}}{2}\right) \frac{\mathbf{v}}{\sqrt{|\kappa|}\|\mathbf{v}\|_{2}}\right)
$
\\
Logarithmic Mapping & 
$\mathbf{log}_{\mathbf{x}}^{\kappa}(\mathbf{y})= \mathbf{x}-\mathbf{y}$
&
$
\mathbf{log}_{\mathbf{x}}^{\kappa}(\mathbf{y})=\frac{2}{\sqrt{|\kappa|} \lambda_{\mathbf{x}}^{\kappa}} \tan _{\kappa}^{-1}\left(\sqrt{|\kappa|}\left\|-\mathbf{x} \oplus_{\kappa} \mathbf{y}\right\|_{2}\right) \frac{-\mathbf{x} \oplus_{\kappa} \mathbf{y}}{\left\|-\mathbf{x} \oplus_{\kappa} \mathbf{y}\right\|_{2}}
$
\\
\hline
Addition & 
$\mathbf{x} \oplus_{\kappa} \mathbf{y}=\mathbf{x} + \mathbf{y}$
&
$
\mathbf{x} \oplus_{\kappa} \mathbf{y}=\frac{\left(1+2 \kappa \mathbf{x}^{T} \mathbf{y}+K\|\mathbf{y}\|^{2}\right) \mathbf{x}+\left(1-\kappa || \mathbf{x}||^{2}\right) \mathbf{y}}{1+2 \kappa \mathbf{x}^{T} \mathbf{y}+\kappa^{2}|| \mathbf{x}||^{2}|| \mathbf{v}||^{2}}
$\\
Scalar-vector Multiplication & 
$r \otimes_{\kappa} \mathbf{x}=r \mathbf{x}$
&
$
r \otimes_{\kappa} \mathbf{x}=\mathbf{exp} _{\mathbf{0}}^{\kappa}\left(r \ \mathbf{log}_{\mathbf{0}}^{\kappa}(\mathbf{x})\right) 
$\\
Matrix-vector Multiplication &
$\mathbf M \otimes_{\kappa} \mathbf{x}=\mathbf M \mathbf{x}$
& 
$
\mathbf M \otimes_{\kappa} \mathbf{x}=\mathbf{exp} _{\mathbf{0}}^{\kappa}\left(\mathbf M \ \mathbf{log}_{\mathbf{0}}^{\kappa}(\mathbf{x})\right) 
$\\
Applying Functions &
$f^{\otimes_{\kappa}}(\mathbf x)=f(\mathbf x)$
&
$
f^{\otimes_{\kappa}}(\mathbf x)=\mathbf{exp} _{\mathbf{0}}^{\kappa}\left(f\left(\mathbf{log}_{\mathbf{0}}^{\kappa}(\mathbf x)\right)\right)
$\\
\hline
\end{tabular}
\label{tab:ops}
\end{table*}

\subsubsection{Intra-Component Attentional Layer:}
This is the building block layer of the proposed mixed-curvature GNN.
In this layer, we update node representations by attentionally aggregating the representations of its neighbors in the constant-curvature component space.
%attentional aggregation within component space
% early diffusion
% $$\mathcal G=\sum\nolimits_{t=i}^{\infty}\theta_t \mathbf G^t$$
%We introduce the attention and aggregation respectively.
As the importance of neighbors is usually different, we introduce the \emph{intra-component attention} to learn the importance of the neighbors.
Specifically, we first lift to the tangent space via $\mathbf{log} _\mathbf{0}^{\kappa}$ and model the importance parameterized by  $\boldsymbol \theta_\text{intra}$ as follows:
\begin{equation}
\resizebox{0.87\hsize}{!}{$
\emph{att}_\text{intra}(\boldsymbol z_i, \boldsymbol z_j)= \sigma \left( \boldsymbol{\theta_\text{intra}}^\top( \mathbf W\mathbf{log} _\mathbf{0}^{\kappa}(\boldsymbol z_i) || \mathbf W \mathbf{log} _\mathbf{0}^{\kappa}(\boldsymbol z_j) ) \right),
$}
\end{equation}
where $\mathbf W$ is the shared weight matrix and $\sigma(\cdot)$ denotes the sigmoid activation.  $\mathbf{log} _\mathbf{0}^{\kappa}$ and $\mathbf{exp} _\mathbf{0}^{\kappa}$ are exponential and logarithmic maps defined in Table \ref{tab:ops}, respectively. Then, we compute the attention weight via softmax:
\begin{equation}
\hat{\mathbf A}_{ij}=\frac{\exp(\emph{att}_\text{intra}(\boldsymbol z_i, \boldsymbol z_j))}{\sum_{k \in \mathcal{N}_{i}} \exp(\emph{att}_\text{intra}(\boldsymbol z_i, \boldsymbol z_k))},
\end{equation}
where $\mathcal N_i$ is neighborhood  node index of node $v_i$ in the graph.
Finally, we add a self-loop to keep its initial information, i.e., we have $\mathbf A=\mathbf I +\hat{\mathbf A}$.

%in the $\kappa$-stereographic model 

Aggregation in traditional Euclidean space is straightforward. 
However, aggregation in hyperbolic or spherical space is challenging as the space is curved.
To bridge this gap, we define the row-wise  \emph{$\kappa$-left-matrix-multiplication} $\boxtimes_{\kappa}$ similar to \cite{BachmannBG20}.
Let rows of $\mathbf{Z}$ hold the vectors in $\kappa$-stereographic model. We have
\begin{equation}
\resizebox{0.77\hsize}{!}{$
\left(\mathbf{A} \boxtimes_{\kappa} \mathbf{Z}\right)_{i \bullet}:= A \otimes_{\kappa} \mathbf{mid}_{\kappa}\left( \{ \mathbf{Z}_{i \bullet}\}_{i=1}^n ; \mathbf{A}_{i \bullet}\right),
$}
\end{equation}
where $A=\sum_{j} \mathbf A_{i j}$, $(\cdot)_{i \bullet}$ denotes the $i^{th}$ row and  $\mathbf{mid}_{\kappa}$ denotes the \emph{midpoint} defined as follows:
\begin{equation}
\resizebox{1.05\hsize}{!}{$
\mathbf{mid}_{\kappa}\left( \{\mathbf{Z}_{i \bullet}\}_{i=1}^n ; \mathbf{A}_{i \bullet}\right)=\frac{1}{2} \otimes_{\kappa} \left(\sum_{l=1}^{n} \frac{\mathbf{A}_{il} \lambda_{\mathbf{Z}_{i \bullet}}^{\kappa}}{\sum_{j=1}^{n} \mathbf{A}_{ij} (\lambda_{\mathbf{Z}_{i \bullet}}^{\kappa}-1)} \mathbf{Z}_{i \bullet}\right),
$}
\end{equation}
 where notation $\lambda_{\mathbf{Z}_{i \bullet}}^{\kappa}$ has been defined in Eq. (\ref{conformalfactor}) before.
\emph{We show that $\kappa$-left-matrix-multiplication $\boxtimes_{\kappa}$ performs attentional aggregation, i.e., \textbf{Theorem 1}.}

%In each component space $\mathcal M_i$
With the attention and aggregation above, we are ready to formulate the unified intra-component layer in component $\mathcal M_i$ of arbitrary curvature $\kappa_i$. % as follows:
Given the input $\mathbf{Z}^{(\ell-1)}_{\mathcal M_i}$ holding embeddings in its rows,  
the $\ell^{th}$ layer outputs:
\begin{equation}
\resizebox{0.75\hsize}{!}{$
\mathbf Z^{(\ell)}_{\mathcal M_i}=\sigma^{\otimes_{\kappa}}\left( \mathbf{A}^{(\ell)}_{\mathcal M_i} \boxtimes_{\kappa}\left(\mathbf{Z}^{(\ell-1)}_{\mathcal M_i} \otimes_{\kappa} \mathbf{W}^{(\ell-1)}_{\mathcal M_i}\right)\right),
$}
\end{equation}
where we define the $\kappa$-right-matrix-multiplication below:
\begin{equation}
\resizebox{1\hsize}{!}{$
\left(\mathbf{Z}^{(\ell-1)}_{\mathcal M_i} \otimes_{\kappa} \mathbf{W}^{(\ell-1)}_{\mathcal M_i}\right)_{i \bullet}:= \mathbf{exp} _\mathbf{0}^{\kappa}\left(\mathbf{log} _\mathbf{0}^{\kappa}\left((\mathbf{Z}^{(\ell-1)}_{\mathcal M_i})_{i \bullet}\right) \mathbf{W}^{(\ell-1)}_{\mathcal M_i}\right).
$}
\end{equation}
% $\sigma^{\otimes_{\kappa}}(\mathbf x)=\mathbf{exp} _{\mathbf{0}}^{\kappa}\left(\sigma \left(\mathbf{log}_{\mathbf{0}}^{\kappa}(\mathbf x)\right)\right)$, 
In particular, we have $\mathbf Z^{(0)}_{\mathcal M_i}$ hold the input features in the $\kappa$-stereographic model.
After stacking $L$ layers, we have the output matrix $\mathbf Z_{\mathcal M_i}=\mathbf Z^{(L)}_{\mathcal M_i}$ hold the constant-curvature component embedding $\boldsymbol z_{\mathcal M_i}$ of component space $\mathcal M_i$. 

\newtheorem*{thm}{Theorem 1 ($\kappa$-left-matrix-multiplication as attentional aggregation)}
\begin{thm}
Let rows of $\mathbf{H}$ hold the encoding $\boldsymbol z_{\mathcal M_i}$, linear transformed by $\mathbf{W}$, 
and $\mathbf{A}$ hold the attentional weights,
the  $\kappa$-left-matrix-multiplication $\mathbf{A} \boxtimes_{\kappa} \mathbf{H}$ performs the attentional aggregation over the rows of $\mathbf{H}$, i.e., 
$(\mathbf{A} \boxtimes_{\kappa} \mathbf{H})_{p \bullet}$ is the linear combination of $\mathbf{H}_{p\bullet}$ with respect to attentional weight $\mathbf{A}_{pq}$,
% \begin{equation}
% (\mathbf{A} \boxtimes^{\kappa} \mathbf{H})_{i \bullet}=\oplus^{\kappa}_{j\in \Psi}(\mathbf{A}_{ij} \otimes^{\kappa} \mathbf{H}_{i \bullet}),
% \end{equation}
where $q$ enumerates the node index in set $\Psi$, $\Psi=p \cup \mathcal N_p$ and $\mathcal N_p$ is the neighborhood node index of $v_p$.
\end{thm}
\begin{proof}
Please refer to the Supplementary Material.
\end{proof}

\subsubsection{Inter-Component Attentional Layer:}
%attentional concatenation among component space
This is the output layer of the proposed mixed-curvature GNN.
In this layer, we perform attentional concatenation to fuse constant-curvature representations across component spaces so as to learn mixed-curvature representations in the product space.

The importance of constant-curvature component space is usually different in constructing the mixed-curvature space.
Thus, we introduce the inter-component attention to learn the importance of component space.
Specifically, we first lift component encodings to the common tangent space, and figure out their centorid by the mean pooling as follows:
\begin{equation}
\boldsymbol \mu=\text{Pooling}\left( \mathbf{log}_{\mathbf 0}^\kappa(\mathbf W_i \otimes_\kappa \boldsymbol z_{\mathcal M_i}) \right),
\end{equation}
where we construct the common tangent space by the linear transformation $\mathbf W_i $ and  $\mathbf{log}_{\mathbf 0}^\kappa$.
Then, we model the importance of a component by the position of the component embedding relative to the centorid,  parameterized by  $\boldsymbol \theta_\text{inter}$,
\begin{equation}
\resizebox{0.89\hsize}{!}{$
\emph{att}_\text{inter}(\boldsymbol z_{\mathcal M_i}, \boldsymbol \mu)=\sigma\left({\boldsymbol \theta_\text{inter}}^{\top}\left(\mathbf{log}_{\mathbf 0}^\kappa(\mathbf W_i \otimes_\kappa \boldsymbol z_{\mathcal M_i}) - \boldsymbol \mu \right)\right),
$}
\end{equation}
 Next, we compute the attention weight of each component via the softmax function as follows:
\begin{equation}
w_i=
\frac{\exp( \emph{att}_\text{inter}(\boldsymbol z_{\mathcal M_i}, \boldsymbol \mu) )}
{\sum_{k=1}^K   \exp( \emph{att}_\text{inter}(\boldsymbol z_{\mathcal M_k}, \boldsymbol \mu)  ) }.
\end{equation}
Finally, with the learnable attentional weights, we perform attentional concatenation and have the output representation, $\boldsymbol z= ||_{i=1}^K (\boldsymbol w_i \otimes_\kappa \boldsymbol z_{\mathcal M_i})$.
Note that, learning representations in the mixed-curvature space not only matches the complex structures of graphs,
but also inherently provides the positive and negative samples of multiple Riemannian views for contrastive learning, which we will discuss in the next part.

\subsection{Dual Contrastive Approach}
With the combinatorial construction of the mixed-curvature space, we propose a novel \emph{dual contrastive approach} of single-view and cross-view contrastive learning for the self-supervisd learning.
%we first reveal three Riemannian views with respect of the sign of curvature by the proposed Riemannian Projector, and then contrast the positive and negative samples via the Riemannian Discriminator $\mathcal D_\kappa$.
To this end, we first design a Riemannian Projector to reveal the hyperbolic, spherical and Euclidean views with respect of the sign of curvature $\kappa$, and then design a Riemannian Discriminator $\mathcal D_\kappa$ to contrast the positive and negative samples.
As shown in Fig. \ref{illu}, 
we contrast the samples in the same Riemannian view (i.e., \emph{single-view contrastive learning})
and concurrently contrast across different Riemannian views (i.e.,  \emph{cross-view contrastive learning}). 
%We illustrate the dual contrastive approach in Fig. x.
We summarize the self-supervised learning process of \text{SelfMGNN} with dual contrastive loss in Algorithm \ref{algo}.

\subsubsection{Riemannian Projector:}
We design the Riemannian projector to reveal different Riemannian views for contrastive learning.
Recall that the mixed-curvature space $\mathcal P$ is a combinatorial construction of $3$ canonical types of component spaces in essence, i.e., $\mathbb H$ ($\kappa<0$), $\mathbb S$ ($\kappa>0$) and $\mathbb E$ ($\kappa=0$), where $\mathbb H$ and $\mathbb S$ can have multiple component spaces with distinct learnable curvatures.
We can fuse component encodings of the same space type and obtain $3$ canonical Riemannian views: hyperbolic  $\mathbf h \in \mathbb H$, spherical $\mathbf s \in \mathbb S$  and Euclidean $\mathbf e \in \mathbb E$.
To this end, we design a map, \emph{RiemannianProjector}: $(\mathbf Z)_{i \bullet} \to [\mathbf h_i \ \ \mathbf e_i \ \ \mathbf s_i] $, 
where $(\mathbf Z)_{i \bullet}$ is the output of mixed-curvature GNN containing all component embeddings.
% of the corresponding type.
%Obtain the projected embedding in each geometric view
Specifically, for each space type, we first project the component embedding $\boldsymbol z_{\mathcal M_i}$ to the corresponding space of standard curvature via $\text{MLP}_\kappa$ layers defined as follows:
\begin{equation}
\text{MLP}_\kappa( \mathbf x)=\sigma^{\otimes_\kappa}( \mathbf b \oplus_\kappa \mathbf M \otimes_\kappa \boldsymbol z_{\mathcal M_i} ),
 \end{equation}
where $\mathbf M$ and  $\mathbf b$ denote the weight matrix and bias, respectively.
Then, we fuse the projected embeddings in account of the importance of component space via the $\mathbf{mid}_{\kappa}$ function:
\begin{equation}
\mathbf v= \mathbf{mid}_{\kappa}\left( \{ \mathbf W_i \otimes_\kappa \boldsymbol z_{\mathcal M_i}\}_{i \in \Omega}; \{{w}_i\}_{i \in \Omega} \right),
 \end{equation}
where $\Omega$ is the component index set of the given type. $\mathbf{W}_i$ is the linear transformation. The importance weight $\boldsymbol{w}_i$ is learned by \emph{inter-component attention}, and  $\mathbf v \in \{\mathbf h, \mathbf e, \mathbf s\}$.

\begin{algorithm}[tb]
        \caption{Self-supervised Learning S\footnotesize{\textsc{elfMGNN}}} 
        % \LinesNumbered
        \KwIn{Graph $ G=(\mathbf G, \mathbf X)$, weight $\gamma$, Congruent Graph Generation Function $\Gamma(\cdot)$ }
        \KwOut{\emph{MixedCurvatureGNN} para., and throw away \emph{RiemannianProjector} para.}
        \While{not converging}{
            %Draw a congruent graph generation function $\Gamma(\cdot)$\;
            \textcolor{cyan}{\emph{//  Views of the original graph $\mathbf G^\alpha$:}}

            Set $\mathbf G^\alpha = \mathbf G$\;
            $\mathbf Z^\alpha = \emph{MixedCurvatureGNN}(\mathbf G^\alpha, \mathbf X; \boldsymbol{\theta}^\alpha)$\;
            $[\mathbf H^\alpha \ \mathbf E^\alpha \ \mathbf S^\alpha] = \emph{RiemannianProjector}(\mathbf Z^\alpha; \boldsymbol{\phi})$\;
            \textcolor{cyan}{\emph{//  Views of the congruent augmentation $\mathbf G^\beta$:}}

           Generate a congruent graph  $\mathbf G^\beta = \Gamma(\mathbf G)$\;
            $\mathbf Z^\beta = \emph{MixedCurvatureGNN}(\mathbf G^\beta, \mathbf X; \boldsymbol{\theta}^\beta)$\;
            $[\mathbf H^\beta \ \mathbf E^\beta \ \mathbf S^\beta] = \emph{RiemannianProjector}(\mathbf Z^\beta; \boldsymbol{\phi})$\;
            \textcolor{cyan}{\emph{// Dual contrastive loss:}}

            \For{each node $v_i$ in $\mathbf G^\alpha$ and $v_j$ to $\mathbf G^\beta$}{
                    \For{Riemannian views $\mathbf x,  \mathbf y \in \{\mathbf h, \mathbf e, \mathbf s\}$}{
                            %Compute Eqs. (\ref{alpha_1}) and (\ref{alpha_2}) with the discriminator $\mathcal D^\kappa(\mathbf x^\alpha, \mathbf x^\beta; \mathbf D_T)$\;
                            Single-view contrastive learning with Eqs. (\ref{alpha_1}) and (\ref{alpha_2})\; 
                            Cross-view contrastive learning with Eqs. (\ref{geo_1}) and (\ref{geo_2})\; 
                    }
            }
            \textcolor{cyan}{\emph{// Update neural network parameters:}}

            Compute gradients of the dual contrastive loss:
            $$
            \nabla_{\boldsymbol{\theta}^\alpha, \ \boldsymbol{\theta}^\beta, \ \boldsymbol{\phi},\ \mathbf D_S, \ \mathbf D_C}\ \ \mathcal J_S + \lambda \mathcal J_C. 
            $$
            }
            \label{algo}
\end{algorithm}

\subsubsection{Riemannian Discriminator:}
Contrasting between positive and negative samples is fundamental for contrastive learning.
However, it is challenging in the Riemannian space, and existing methods, to our knowledge,  cannot be applied to Riemannian spaces due to the intrinsic difference in the geometry.
To bridge this gap, we design a novel Riemannian Discriminator to scores the agreement between positive and negative samples.
The main idea is that we lift the samples to the common tangent space, 
and evaluate the agreement score in the tangent space.
Specifically, we utilize the bilinear form to evaluate the agreement.
Given two Riemannian views $\mathbf x$ and $\mathbf y$ of a node, $\mathbf x, \mathbf y \in \{\mathbf h, \mathbf e, \mathbf s\}$, we give the formulation parameterized by the matrix $\mathbf D$ as follows:
\begin{equation}
\mathcal D_\kappa(\mathbf x, \mathbf y)=\left(\mathbf{log}_{\mathbf 0}^{\kappa_{\mathbf x}}(\mathbf x)\right)^\top\mathbf D \left(\mathbf{log}_{\mathbf 0}^{\kappa_{\mathbf y}}(\mathbf y)\right),
\end{equation}
where we construct the common tangent space via $\mathbf{log}_{\mathbf 0}^{\kappa_{\mathbf x}}$,
and $\kappa_{\mathbf x}$ is the curvature of the corresponding view.

\subsubsection{Single-view Contrastive Learning:}
%we contrast positive and negative samples in the same Riemannian view, \emph{single-view contrastive learning}
\textsc{SelfMGNN} employs the single-view contrastive learning in each Riemannian view of the mixed-curvature space.
% Specifically, for the graph $\mathbf G^\alpha$, 
% we first include a congruent counterpart $\mathbf G^\beta$ similar to \cite{ChenK0H20,HassaniA20}.
Specifically, we first include a congruent augmentation $\mathbf G^\beta$ similar to \citet{ChenK0H20,HassaniA20}.
Then, we introduce a contrastive discrimination task for a given Riemannian view:
for a sample in  $\mathbf G^\alpha$, we aims to discriminate the positive sample from negative samples in the congruent counterpart $\mathbf G^\beta$.
Here, we use superscript $\alpha$ and $\beta$ to distinguish notations of the graph and its congruent augmentation.
%maximize the MI between two views by contrasting node representations of one component space with another component space
% $\sum_{i=1}^{|V|}\sum_{\mathbf x, \mathbf y \in \{\mathbf h, \mathbf e, \mathbf s\}}\text{MI}(\mathbf x_i^\alpha, \mathbf x_i^\beta) - 
% \mathbb{I}\left[ \mathbf x \neq \mathbf y\right]
% \text{MI}(\mathbf x_i^\alpha, \mathbf y_i^\beta)$
We formulate the InfoNCE loss \cite{abs-1807-03748} as follows:
\begin{equation}
\mathcal L_S(\alpha, \beta)=-\log \frac{\exp \mathcal D_\kappa(\mathbf x_i^\alpha, \mathbf x_i^\beta)}{\sum_{j=1}^{|V|}\mathbb I\{i \neq j\}\exp \mathcal D_\kappa(\mathbf x_i^\alpha, \mathbf x_j^\beta)},
\label{alpha_1}
\end{equation}
where $\mathbf x_i^\beta$ and $\mathbf x_j^\beta$ are the positive sample and negative samples of $v_i$ in $\mathbf G^\alpha$, respectively. 
$\mathbb I\{ \cdot \} \in \{0, 1\}$ is an indicator function who will return $1$ iff the condition $(\cdot)$ is true ($i \neq j$ in this case).
We utilize the Riemannian discriminator $\mathcal D_\kappa(\cdot, \cdot)$ to evaluate the agreement between the samples.

In the single-view contrastive learning, 
for each Riemannian view, 
we contrast between $\mathbf G^\alpha$ and its congruent augmentation $\mathbf G^\beta$, and vice versa. 
Formally, we have the single-view contrastive loss as follows:
%augmentation
\begin{equation}
\mathcal J_S = \sum\nolimits_{i=1}^{|V|}\sum\nolimits_{\mathbf x \in \{\mathbf h, \mathbf e, \mathbf s\}}\left(\mathcal L_S(\alpha, \beta)+ \mathcal L_S(\beta, \alpha)\right).
\label{alpha_2}
\end{equation}

 \begin{table*}
  \scriptsize
    \centering
          \caption{The summary of AUC (\%) for link prediction (LP) and classification accuracy (\%) for node classification (NC) on Citeseer, Cora, Pubmed, Amazon and USA datasets. The highest scores are in \textbf{bold}, and the second \textcolor{blue}{blue}.}
    \begin{tabular}{p{0.22cm}<{\centering} p{1.75cm}<{\centering}|p{0.97cm}<{\centering} p{1.15cm}<{\centering} |p{0.97cm}<{\centering} p{1.15cm}<{\centering} |p{0.97cm}<{\centering} p{1.15cm}<{\centering}| p{0.97cm}<{\centering} p{1.15cm}<{\centering}| p{0.97cm}<{\centering} p{1.1cm}<{\centering}}
      \toprule
  \multicolumn{2}{c|}{}        & \multicolumn{2}{c|}{ \footnotesize{\textbf{Citeseer}}} &  \multicolumn{2}{c|}{ \footnotesize{\textbf{Cora}} } &  \multicolumn{2}{c|}{ \footnotesize{\textbf{Pubmed}} } &  \multicolumn{2}{c|}{ \footnotesize{\textbf{Amazon}} } &  \multicolumn{2}{c}{ \footnotesize{\textbf{Airport}} }\\
  \multicolumn{2}{c|}{ \footnotesize{\textbf{Method}}  }& \footnotesize{LP}& \footnotesize{NC} & \footnotesize{LP} & \footnotesize{NC}& \footnotesize{LP}& \footnotesize{NC} & \footnotesize{LP} & \footnotesize{NC} & \footnotesize{LP} & \footnotesize{NC}\\
     \toprule
       \multirow{6}{*}{\rotatebox{90}{\footnotesize{Euclidean}} } 
                         &\footnotesize{GCN}   
&  $ 93.6(0.7)$   &  $  70.2(0.8)$   &  $ 91.4(0.7)$  & $  81.3(0.3)$   & $93.0(0.6)$    &  $ 78.8(0.2)$    &  $92.9(0.9)$  & $71.2(1.1)$ &  $ 90.5(0.4)$ &  $ 50.8(0.9)$ \\
                &  \footnotesize{GraphSage}
&  $ 87.2(0.9)$   & $  68.2(1.1)$   &  $  88.7(0.6)$  & $  78.1(0.8)$    &$ 87.7(0.4)$    &  $77.5(0.3)$   &  $91.8(0.5)$  & $72.9(1.6)$ &  $ 85.6(1.1)$ &  $47.8(0.8)$ \\
                        &\footnotesize{GAT}     
&  $ 92.9(0.7)$   &  $72.0(0.7)$     & $  93.4(0.4)$ &  $  82.1(0.7)$    & $92.6(0.3)$    &  $ 77.1(0.7)$   &  $93.9(0.6)$  & $72.6(0.8)$ &  $91.4(0.6)$  &  $ 49.3(0.7)$\\
                           & \footnotesize{DGI }
&  $ 92.7(0.5)$   &  $ 71.3(0.7)$    & $  91.8(0.5)$ & $  81.4(0.6)$    & $92.8(0.7)$    &  $ 76.6(0.6)$   &  $93.5(0.4)$   & $72.2(0.3)$ & $92.5(0.8)$   &   $ 50.1(0.5)$ \\
                    &  \footnotesize{MVGRL }
& $  94.8(0.3)$   &  $ 72.1(0.8)$    &$  93.2(0.7)$   & $  82.7(0.7)$    & $95.9(0.2)$    &  $ 78.9(0.3)$   &  $96.2(0.5)$  & $74.0(1.0)$ & $95.1(0.3)$ & $\blue{52.1}(1.0)$\\
                        &   \footnotesize{GMI}   
&$  95.0(0.6)$& $\blue{72.5}(0.3)$&$\blue{93.9}(0.3)$&$81.8(0.2)$&$96.5(0.8)$ &$\blue{79.0}(0.2)$&$96.8(0.7)$   & $74.5(0.9)$ & $94.7(0.5)$   &   $ 51.9(0.7)$ \\
       \midrule
       \multirow{5}{*}{\rotatebox{90}{\footnotesize{Riemannian}}}
                      &  \footnotesize{HGCN} 
&  $  94.6(0.4)$  & $71.7(0.5)$  &  $ 93.2(0.1)$  &   $ 81.5(0.6)$  &  $  96.2(0.2)$   &   $ 78.5(0.4)$   &  $ 96.7(0.9)$&$\blue{75.2}(1.3)$& $93.6(0.3)$    &  $  51.2(0.6)$ \\
                          &  \footnotesize{HAT }
&  $  93.7(0.5)$  & $72.2(0.6)$   & $93.0(0.5)$   &   $ 83.1(0.7)$ &  $  96.3(0.3)$     &  $78.6(0.7)$   & $  96.9(1.1)$   &$  74.1(1.0)$  & $  93.9(0.6)$    &   $  51.3(1.0)$ \\
                       & \footnotesize{ LGCN }
&$\blue{95.5}(0.5)$&$72.1(0.7)$& $93.7(0.5)$  &$\blue{83.3}(0.9)$&$\blue{96.6}(0.2)$&$78.6(1.0)$ &$\mathbf{97.5}(0.9)$ & $75.1(1.1)$&$\blue{96.4}(0.2)$&$52.0(0.9)$ \\
          &  \footnotesize{$\kappa$-GCN}
&  $  93.8(0.7)$  & $71.2(0.5)$ & $ 92.8(0.8)$   &   $ 81.6(0.7)$  &  $  95.0(0.3)$     & $ 78.7(0.6)$   &  $  94.8(0.6)$ & $  72.4(1.5)$  & $  93.5(0.7)$     &  $  50.9(1.2)$ \\
&\footnotesize{\textbf{\textsc{SelfMGNN}}}
 &  $\mathbf{96.9}(0.3)$ & $\mathbf{73.1}(0.9)$ & $\mathbf{94.6}(0.6)$ & $\mathbf{83.8}(0.8)$ & $\mathbf{97.3}(0.2)$ & $\mathbf{79.6}(0.5)$ 
 & $\mathbf{97.5}(1.0)$ & $\mathbf{75.3}(0.8)$  & $\mathbf{96.9}(0.5)$ & $\mathbf{52.7}(0.7)$  \\
      \bottomrule
    \end{tabular} 
        \label{results}
  \end{table*}

\subsubsection{Cross-view Contrastive Learning:}
\textsc{SelfMGNN} further employs a novel cross-view contrastive learning.
The novelty lies in that our design in essence enjoys the multi-view nature of the mixed-curvature space, i.e., we exploit the multiple Riemannian views of the mixed-curvature space, and contrast across different views.
Specifically, we formulate the contrastive discrimination task as follows:
for a given Riemannian view of  $\mathbf G^\alpha$, we aim to discriminate the given view from the other Riemannian views of  $\mathbf G^\beta$.
We formulate the InfoNCE loss as follows:
\begin{equation}
\mathcal L_C(\alpha, \beta)=-\log \frac{\exp \mathcal D_\kappa(\mathbf x_i^\alpha, \mathbf x_i^\beta)}{\sum_{\mathbf x}\mathbb I\{\mathbf x \neq \mathbf y\} \exp \mathcal D_\kappa(\mathbf x_i^\alpha, \mathbf y_i^\beta)},
\label{geo_1}
\end{equation}
where $\mathbb I\{\mathbf x \neq \mathbf y\}$ is to select the embeddings of different Riemannian views. % in order to construct the negative samples.
Similarly, we contrast between $\mathbf G^\alpha$ and $\mathbf G^\beta$ and vice versa, and have the cross-view contrastive loss:
\begin{equation}
\mathcal J_C = \sum\nolimits_{i=1}^{|V|}\sum\nolimits_{\mathbf x \in \{\mathbf h, \mathbf e, \mathbf s\}}\left(\mathcal L_C(\alpha, \beta)+ \mathcal L_C(\beta, \alpha)\right).
\label{geo_2}
\end{equation}

\subsubsection{Dual Contrastive Loss:} In \text{SelfMGNN}, we integrate the single-view and cross-view contrastive learning, and formulate the dual contrastive loss as follows:
\begin{equation}
\mathcal J_{self} = \mathcal J_S + \gamma \mathcal J_C,
\end{equation}
where $\gamma$ is the balance weight. % to  the importance of cross-view contrastive learning.  
The benefit of dual contrastive loss is that 
we can contrast the samples in the same Riemannian view (single-view) and contrast across different Riemannian views (cross-view), % in the mixed-curvature space.
comprehensively leveraging the rich information in the mixed-curvature space to encode the graph structure.
Finally, \text{SelfMGNN} learns representations in the mixed-curvature Riemannian space capturing the complex structures of graphs without labels.

%% file: experiments.tex
%!TEX root = ./SelfMGNN.tex

\section{Experiments}
In this section, we evaluate  \textsc{SelfMGNN} with the link prediction and node classification tasks against $10$ strong baselines on $5$ benchmark datasets. 
%We repeat each experiment $10$ times and report the mean with the standard deviations.
We report the mean with the standard deviations of $10$ independent runs for each model to achieve fair comparisons.

\subsection{Experimental Setups}
\noindent\textbf{Datasets:} We utilize $5$ benchmark datasets, i.e., 
the widely-used 
\textbf{Citeseer}, 
\textbf{Cora},
and
\textbf{Pubmed} \cite{kipf2016semi,VelickovicFHLBH19}, 
and the latest
\textbf{Amazon}
and
\textbf{Airport} \cite{ZhangWSLS21}.
%We list the statistics in Supplementary Material.

\noindent\textbf{Euclidean Baselines:} %We compare the proposed \textsc{SelfMGNN} with the state-of-the-art baselines of two categories: % of constant-curvature space and self-supervised learning.
i) \emph{Supervised Models}: 
\textbf{GCN} \cite{kipf2016semi},
\textbf{GraphSage} \cite{hamilton2017inductive},
\textbf{GAT} \cite{velickovic2018graph}.
ii) \emph{Self-supervised Models}: 
\textbf{DGI} \cite{VelickovicFHLBH19},
\textbf{MVGRL} \cite{HassaniA20},
\textbf{GMI} \cite{PengHLZRXH20}.

\noindent\textbf{Riemannian Baselines:}
i) \emph{Supervised Models}: 
\textbf{HGCN} \cite{HGCN}, 
\textbf{HAT} \cite{HAN}
and
\textbf{LGCN} \cite{ZhangWSLS21} for hyperbolic space; 
$\boldsymbol \kappa$-\textbf{GCN} \cite{BachmannBG20} with positive $\kappa$ for spherical space.
ii)  
\emph{Self-supervised Models}: There is no self-supervised Riemannian models in the literature, and thus we propose \textsc{SelfMGNN} to fill this gap.

%\emph{Refer Supplementary Material for details.}

\subsection{Implementation Details}

%$\kappa$-stereographic model
% \noindent\textbf{Euclidean input:} The input feature $\mathbf x$ is Euclidean by default. 
% In this case, we map it to the Riemannian space to feed into \textsc{SelfMGNN}. 
% Specifically, the map $\rho: \mathbb R^d \to \mathcal M^d_\kappa$ is defined as 
% $\rho(\mathbf x)=( F_\kappa
% |\kappa|^{\frac{1}{2}}
% \sinh(
% |\kappa|^{-\frac{1}{2}}
% ||\mathbf x||_2
% )
% )\mathbf x$,
% where projection factor $F$ is given as follows:
%    \vspace{-0.09in}
% \begin{equation}
% F_\kappa=(1+\kappa\cosh(|\kappa|^{-\frac{1}{2}}||\mathbf x||_2))||\mathbf x||_2.
%    \vspace{-0.06in}
% \end{equation}
% \emph{We derive the map  $\rho(\mathbf x)$ in the Supplementary Material.}

\noindent\textbf{Congruent graph:} 
As suggested by \citet{HassaniA20}, we opt for the diffusion to generate a congruent augmentation.
%Generate congruent views with diffusion matrices
% \begin{equation}
% \Gamma(\mathbf G)=\alpha\left(\mathbf{I}_{n}-(1-\alpha) \mathbf{D}^{-\frac{1}{2}} \mathbf{G D}^{-\frac{1}{2}}\right)^{-1}
% \end{equation}
Specifically, given an adjacency matrix $\mathbf G^\alpha$, we use the congruent graph generation function \emph{$\Gamma(\cdot)$} to obtain a diffusion matrix $\mathbf G^\beta$  and treat it as the adjacency matrix of the congruent augmentation. 
% In practice, we utilize the Personalized PageRank diffusion which has a closed-form formulation: 
%    \vspace{-0.05in}
% \begin{equation}
% \Gamma(\mathbf G^\alpha)=\alpha\left(\mathbf{I}_{n}-(1-\alpha) \mathbf{D}^{-\frac{1}{2}} \mathbf{G^\alpha D}^{-\frac{1}{2}}\right)^{-1},
%    \vspace{-0.05in}
% \end{equation}
% where $\alpha$ denotes teleport probability in the diffusion. 
% Note that, 
The diffusion is computed once via fast approximated and sparsified method \cite{KlicperaWG19}.

\noindent\textbf{Signature:} The mixed-curvature space is parameterized by the signature, i.e., space type, curvature and dimensionality of the component spaces.
The space type of component $\mathcal M_i$ can be hyperbolic $\mathbb H$, spherical $\mathbb S$ or Euclidean $\mathbb E$, and we utilize the combination of them to cover the mixed and complicated graph structures.
The dimensionality $d_{\mathcal M_i}$ is a hyperparameter.
The curvature $\kappa_{\mathcal M_i}$ is a learnable parameters as our loss is differentiable with respect to the curvature. 

\noindent\textbf{Learning manner:} Similar to \citet{VelickovicFHLBH19}, self-supervised models first learn representations without labels, and then were evaluated by specific learning task, which is performed by directly using these representations to train and test for learning tasks.
Supervised models were trained and tested by following \citet{HGCN}.
Please refer to the Supplementary Material for further experimental details.

%We learn the curvatures using gradient based optimization.

%We stack the attentive aggregation layer twice in the experiment.

 \begin{table}
  \scriptsize
    \centering
          \caption{Ablation study of \textsc{SelfMGNN} for node classification task in classification accuracy ($\%$).}
    \begin{tabular}{p{0.05cm}<{\centering} p{2.7cm}<{\centering}|p{1.2cm}<{\centering} p{1.2cm}<{\centering} p{1.2cm}<{\centering} }
      \toprule
\multicolumn{2}{c|}{\footnotesize{\textbf{Variants}}} & \footnotesize{Citesser }& \footnotesize{Core} & \footnotesize{Pubmed}\\[0.5pt]
\toprule
\multirow{3}{*}{\rotatebox{90}{\footnotesize{CCS} } }
                                      &  \footnotesize{$\mathbb H^{24}$}                               &       $  72.2(0.7)$      &    $ 82.1(0.4)$      &  $  78.6(0.3)$        \\
                                      &  \footnotesize{$\mathbb S^{24}$}                                &       $  70.5(0.8)$      &    $ 82.3(0.5)$      &  $  77.5(0.4)$           \\
                                      & \footnotesize{$\mathbb E^{24}$}                                 &       $  71.8(1.1)$      &    $ 81.0(0.7)$      &  $  77.3(0.8)$       \\
\midrule
\multirow{3}{*}{\rotatebox{90}{\footnotesize{Single} }  }
&  \footnotesize{ $\mathbb H^8\times\mathbb S^8 \times \mathbb E^8  $}          &       $ 72.6 (0.3)$      &    $  82.7(0.8)$      &  $  78.9(0.9)$        \\
&  \footnotesize{$(\mathbb H^4)^2\times (\mathbb S^4)^2\times\mathbb E^8 $}&       $ 72.8(0.6)$      &    $  83.1(0.6)$      &  $  79.2(0.2)$       \\
&  \footnotesize{$(\mathbb H^2)^4\times(\mathbb S^2)^4\times\mathbb E^8 $} &       $ 72.9(0.2)$      &    $  83.5(0.5)$      &  $  79.3(0.5)$            \\
\midrule
\multirow{3}{*}{\rotatebox{90}{\footnotesize{Ours} } }
& \footnotesize{ $\mathbb H^8\times\mathbb S^8 \times \mathbb E^8  $ }          &       $ 72.8(1.0)$       &    $  83.3(0.9)$      &  $  79.2(0.6)$           \\
& \footnotesize{ $(\mathbb H^4)^2\times (\mathbb S^4)^2\times\mathbb E^8 $}&  $\blue{73.1}(0.9)$ &$\blue{83.8}(0.5)$ &$\blue{79.6}(0.7)$         \\
& \footnotesize{ $(\mathbb H^2)^4\times(\mathbb S^2)^4\times\mathbb E^8 $} &$\mathbf{73.3}(0.5)$ & $\mathbf{84.1}(0.8)$  &$\mathbf{79.9}(1.1)$           \\
      \bottomrule
    \end{tabular} 
        \label{ablation}
  \end{table}

\subsection{Link Prediction}

%The task of link prediction is to predict the probability of two nodes being connected.  is a generalized sigmoid function, and
For link perdition, we utilize the Fermi-Dirac decoder with distance function to define the probability based on model outputs $\boldsymbol z$.
Formally, we have the probability as follows: % representations. 
%we have Formally, 
\begin{equation}
\resizebox{0.888\hsize}{!}{$
p((i,j) \in E| \boldsymbol z_i, \boldsymbol z_j)=\left(\exp\left((d_\mathcal M(\boldsymbol z_i, \boldsymbol z_j)^2-r)/t\right)+1\right)^{-1},
$}
\end{equation}
where $r$, $t$ are hyperparameters. 
For each method, $d_\mathcal M(\boldsymbol z_i, $ $\boldsymbol z_j)$ is the distance function of corresponding representation space, e.g., $||\boldsymbol z_i - \boldsymbol z_j||_2$ for Euclidean models, and we have
\begin{equation}
\resizebox{0.72\hsize}{!}{$
d_\mathcal P(\boldsymbol z_i, \boldsymbol z_j)^2=\sum\nolimits_{l=1}^K d^{\ \kappa_l}_{\mathcal M_l}\left(({\boldsymbol z_i})_{\mathcal M_l}, ({\boldsymbol z_j})_{\mathcal M_l} \right)^2,
$}
\label{dist}
\end{equation}
for \textsc{SelfMGNN}.
We utilize AUC as the evaluation metric and summarize the performance in Table \ref{results}.
We set output dimensionality to be $24$ for all models for fair comparisons.
Table \ref{results} shows that \textsc{SelfMGNN} outperforms the self-supervised models in Euclidean space consistently since it better matches the mixed structures of graphs with the mixed-curvature space.
\textsc{SelfMGNN} achieves competitive and even better results with the supervised Riemannian baselines. 
The reason lies in that we leverage dual contrastive approach to exploit the rich information of data themselves in the mixed-curvature Riemannian space.

\subsection{Node Classification}
% %\textsc{SelfMGNN} output node representations in the mixed-curvature space. 
% %In this case, our task is to classify nodes based on node representations.
% Most of existing classifiers work with Euclidean spaces, and cannot apply to Riemannian spaces.
% Recently, \cite{HNN} presents a hyperbolic classifier, but it still cannot work with mixed-curvature spaces. % in general.
% %To bridge this gap, similar to the study \cite{HGNN}, we employ an Euclidean classification space.
For node classification, we first discuss the classifier as none of existing classifiers, to our knowledge, can work with mixed-curvature spaces.
To bridge this gap, inspired by \cite{HGNN}, for Riemannian models, we introduce the Euclidean transformation to generate an encoding, summarizing the structure of node representations.
%we utilize the Euclidean encodings which summarize the structure of node representations for classification, similar to \cite{HGNN}.
Specifically, we first introduce a set of centroids $\{\boldsymbol \mu_1,  \cdots, \boldsymbol \mu_C\}$,
where $\boldsymbol \mu_c$ is the centroid in Riemannian space learned jointly with the learning model. %using backpropagation.
% Then, we transform the output representation $\boldsymbol z_j \in \mathcal M$ into an Euclidean encoding $\boldsymbol \xi\in \mathbb R^C$, 
% which summarizes the position of $\boldsymbol z_i $ relative to the centroids, i.e., $\boldsymbol \xi=\left(\xi_{1j}, \ldots, \xi_{Cj}\right) $ and $\xi_{ij}=d_{\mathcal M}(\boldsymbol \mu_i, \boldsymbol z_j)$.
Then, for output representation $\boldsymbol z_j \in \mathcal M$, its encoding is defined as $\boldsymbol \xi=\left(\xi_{1j}, \ldots, \xi_{Cj}\right) $, where $\xi_{ij}=d_{\mathcal M}(\boldsymbol \mu_i, \boldsymbol z_j)$, summarizing the position of $\boldsymbol z_i $ relative to the centroids.
Now, we are ready to use logistic regression for node classification and the likelihood is 
\begin{equation}
p( y | \boldsymbol h )=\text{sigmoid}(\mathbf w_C^\top \boldsymbol h),
\end{equation}
where $\mathbf w_C \in \mathbb R^{|C|}$ is the weight matrix, and $y$ is the label.
$\boldsymbol h=\boldsymbol \xi$ for Riemannian models and $\boldsymbol h$ is the output of Euclidean ones.
We utilize classification accuracy \cite{kipf2016semi} as the evaluation metric and summarize the performance in Table \ref{results}.
\textsc{SelfMGNN} achieves the best results on all the datasets.

 \begin{table}
  \scriptsize
    \centering
          \caption{Learning results of the mixed-curvature space on the datasets — curvature (\textcolor{magenta}{weight}) of  each component space.}
    \begin{tabular}{ p{0.93cm}<{\centering}|p{1.1cm}<{\centering} p{1.1cm}<{\centering} p{1.1cm}<{\centering} p{1.1cm}<{\centering} p{0.65cm}<{\centering}}
      \toprule
\footnotesize{\textbf{Dataset}}&   \footnotesize{$\mathbb H^4$}    &  \footnotesize{$\mathbb H^4$} &     \footnotesize{$\mathbb S^4$}    &  \footnotesize{$\mathbb S^4$} & \footnotesize{$\mathbb E^8$}         \\
\toprule
\footnotesize{Citeseer}&$-0.67(\textcolor{magenta}{0.29})$    &  $-0.58(\textcolor{magenta}{0.19})$   &  $  +0.82 (\textcolor{magenta}{0.21})$   &  $  +2.72 (\textcolor{magenta}{0.13})$ & $ 0 (\textcolor{magenta}{0.18})$     \\
\footnotesize{Cora}      &$-0.90  (\textcolor{magenta}{0.18})$    &  $-1.31  (\textcolor{magenta}{0.25})$   & $  +0.76 (\textcolor{magenta}{0.28})$    & $+0.19 (\textcolor{magenta}{0.08})$ &  $ 0 (\textcolor{magenta}{0.21})$     \\
\footnotesize{Pubmed}&$-1.12  (\textcolor{magenta}{0.26})$     &   $-0.79  (\textcolor{magenta}{0.34})$   &  $ +0.59 (\textcolor{magenta}{0.16})$  & $+1.05 (\textcolor{magenta}{0.15})$  &$ 0 (\textcolor{magenta}{0.09})$    \\
\footnotesize{Amazon}&$-0.78 (\textcolor{magenta}{0.11})$      &  $-1.02  (\textcolor{magenta}{0.48})$   &  $  +1.13 (\textcolor{magenta}{0.05})$  & $+2.24 (\textcolor{magenta}{0.24})$ &$ 0 (\textcolor{magenta}{0.12})$   \\
\footnotesize{Airport}  &$-1.26 (\textcolor{magenta}{0.30})$     &  $-2.15  (\textcolor{magenta}{0.17})$   &  $ +1.85 (\textcolor{magenta}{0.20})$   & $+0.67 (\textcolor{magenta}{0.18})$  & $ 0 (\textcolor{magenta}{0.15})$  \\
      \bottomrule
    \end{tabular} 
        \label{component}
  \end{table}

\subsection{Ablation Study} 
We give the ablation study on the importance of i)  mixed-curvature space (MCS) and 
%as well as the differences between different constant-curvature space, e.g., $\mathbb H$ v.s. $\mathbb E$. 
ii) cross-view contrastive learning.
To this end, we include two kinds of variants: CCS and Single, the degenerated \textsc{SelfMGNN}s without some functional module.
CCS variants work without Cartesian product, and thus are learned by the single-view contrastive loss in a constant-curvature space. 
e.g., $\mathbb H^{24}$ is the variant in the hyperbolic space, where the superscript is the dimensionality.
Single variants work with the mixed-curvature space, but disable the cross-view contrastive learning.
The ours are the proposed \textsc{SelfMGNN}s.
A specific instantiation is denoted by Cartesian product, e.g., 
we use $(\mathbb H^4)^2\times(\mathbb S^4)^2\times\mathbb E^8$ as default, 
whose mixed-curvature space is constructed by Cartesian product of  $2$ $\mathbb H^4$,  $2$ $\mathbb S^4$ and $1$ $\mathbb E^8$ component spaces. 

We show the classification accuracy of these variants in Table \ref{ablation}, and we find that:
i) Mixed-curvature variant with single or dual contrastive learning outperforms its CCS counterpart. The reason lies in that the mixed-curvature space is more flexible than a constant-curvature space to cover the complicated graph structures.
ii) Disabling cross-view contrastive learning decreases the performance as the cross-view contrasting further unleashes the rich information of data in the mixed-curvature space. 
iii) Allowing  more than one hyperbolic and spherical spaces can also improve performance as $(\mathbb H^4)^2\times (\mathbb S^4)^2\times\mathbb E^8$
and  $(\mathbb H^2)^4\times (\mathbb S^2)^4\times\mathbb E^8$ both outperform $\mathbb H^8 \times\mathbb S^8\times\mathbb E^8$.

%we list each component space of  $\mathbb H^8\times\mathbb S^8\times\mathbb E^8$ for node classification in Table \ref{component}.
Furthermore, we discuss the curvatures of the datasets. 
We report the learned curvatures and weights of each component space for the real-world datasets in Table \ref{component}.
%Different datasets present different curvature distributions. Hence, the models with fixed curvature is limited.
As shown in Table \ref{component}, component spaces of the same space type are learned with different curvatures, showing that the curvatures over different hierarchical or cyclical regions can still be different.
$(\mathbb H^4)^2\times (\mathbb S^4)^2\times\mathbb E^8$ has $2$ component spaces for hyperbolic (spherical) geometry, 
%which includes the curvatures of $\mathbb H^8\times\mathbb S^8\times\mathbb E^8$.
and allowing multiple hyperbolic (spherical) components enables us to cover a wider range of curvatures of the graph, better matching the graph structures.
This also explains why $(\mathbb H^4)^2\times (\mathbb S^4)^2\times\mathbb E^8$ outperforms  $\mathbb H^8 \times\mathbb S^8\times\mathbb E^8$ in Table \ref{ablation}.
With learnable curvatures and weights of each component, \textsc{SelfMGNN} matches the mixed and complicated graph structures, and learns more promising graph representations. 
%This shows that graph structures are usually heterogeneous. 

%% file: relatedwork.tex
%!TEX root = ./SelfMGNN.tex

\section{Related Work}

%We briefly discuss the related work as follows:
%Our work is related to Riemannian representation learning and contrastive learning. We briefly discuss them as follows:
\textsc{SelfMGNN} learns representations via a mixed-curvature graph neural network with the dual contrastive approach. 
Here, we briefly discuss the related work on the \emph{graph neural network} and \emph{contrastive learning}:
% \noindent\textbf{Graph Neural Network (GNNs)}
% emerge as the state-of-the-art representation learning methods on graphs.
% Existing methods in the literature belong to two major categories: spectral methods and message passing methods.
% The former studies the spectral domain of graphs to learn the representations \cite{BrunaZSL13,kipf2016semi},
% while the latter builds message passing architectures to aggregate neighbors’ features \cite{velickovic2018graph,hamilton2017inductive,ma2019Disentangled}.
% A detailed survey can be found in \cite{WuPCLZY21}.
% Our work attempts to address the limitation of prior GNNs in terms of representation space and learning paradigm.

\subsection{Graph Neural Network}
Graph Neural Networks (GNNs) achieve the state-of-the-art results in graph analysis tasks \cite{DouL0DPY20,pengh2021tois}.
Recently, a few attempts propose to marry GNN and Riemannian representation learning
\cite{HAN,mathieu2019continuous,monath2019gradient}
as graphs are non-Euclidean inherently \cite{krioukov2010hyperbolic}.
In hyperbolic space, \citet{nickel2017poincare,suzuki2019hyperbolic} introduce shallow models,
while \citet{HGNN,HGCN,ZhangWSLS21} formulate deep models (GNNs).
Furthermore, \citet{Sun21hvgnn} generalize hyperbolic GNN to dynamic graphs with insights in temporal analysis. 
\citet{fuxc2020icdm} study the curvature exploration.
\citet{Sun20jointAlign,Wang0Z20} propose promising methods to apply the hyperbolic geometry to network alignment.

Beyond hyperbolic space,
\citet{CruceruBG21} studies the matrix manifold of Riemannian spaces,
and
\citet{BachmannBG20} generalizes GCN to arbitrary constant-curvature spaces.
%\textbf{Constant-curvature space} models work well on the particular kinds of structure they were designed for.
To generalize representation learning, \citet{GuSGR19}  propose to learn in the mixed-curvature space.
\citet{SkopekGB20} introduce the mixed-curvature VAE.
Recently, \citet{WangWSWNAXYC21} model the knowledge graph triples in mixed-curvature space specifically with the supervised manner. 
Distinguishing from these studies, we propose the first self-supervised mixed-curvature model, allowing multiple hyperbolic (spherical) subspaces each with distinct curvatures.

\subsection{Contrastive Learning}
Contrastive Learning is an attractive self-supervised learning method that learns representations by contrasting positive and negative pairs \cite{ChenK0H20}. 
Here, we discuss the contrastive learning methods on graphs.
Specifically,
%DGI \cite{VelickovicFHLBH19} builds local patches and global summary pairs, and contrast via Infomax theory,
\citet{VelickovicFHLBH19} contrast patch-summary pairs via infomax theory.
\citet{HassaniA20} leverage multiple views for contrastive learning.
\citet{QiuCDZYDWT20} formulate a general framework for pre-training.
\citet{WanPY021} incorporate the generative learning concurrently.
\citet{PengHLZRXH20} explore the graph-specific infomax for contrastive learning.
\citet{pmlr-v139-xu21g} aim to learn graph level representations.
\citet{ParkK0Y20,abs-2105-09111} consider the heterogeneous graphs.
To the best of our knowledge, 
existing methods cannot apply to Riemannian spaces % due to the intrinsic difference in the geometry,
and we bridge this gap in this paper.

%  \begin{table}
%   \scriptsize
%     \centering
%           \caption{\textsc{SelfMGNN} ($\mathbb H^8\times\mathbb S^8\times\mathbb E^8$): curvature (\textcolor{magenta}{attentional weight}) for node classification on datasets.}
%             \vspace{-0.1in}
%     \begin{tabular}{ p{1.5cm}<{\centering}|p{1.7cm}<{\centering} p{1.7cm}<{\centering} p{1.7cm}<{\centering} }
%       \toprule
%   \footnotesize{\textbf{Component}} &   \footnotesize{Hyperbolic}       &    \footnotesize{Euclidean}    & \footnotesize{Spherical}        \\
% \toprule
% \footnotesize{Citeseer}&       $  -0.67 \ (\textcolor{magenta}{0.39})$      &    $ 0\ (\textcolor{magenta}{0.28})$     &  $  +0.82\ (\textcolor{magenta}{0.33})$     \\
% \footnotesize{Cora}      &       $  -0.90\  (\textcolor{magenta}{0.48})$      &    $ 0\ (\textcolor{magenta}{0.24})$     & $  +0.76\ (\textcolor{magenta}{0.28})$       \\
% \footnotesize{Pubmed}&       $  -1.12\  (\textcolor{magenta}{0.56})$      &    $ 0\ (\textcolor{magenta}{0.25})$      &  $ +0.59\ (\textcolor{magenta}{0.19})$      \\
% \footnotesize{Amazon}&       $  -0.78\ (\textcolor{magenta}{0.72})$      &    $ 0\ (\textcolor{magenta}{0.13})$      &  $  +1.13\ (\textcolor{magenta}{0.15})$     \\
% \footnotesize{Airport}  &       $  -1.26\ (\textcolor{magenta}{0.67})$      &    $ 0\ (\textcolor{magenta}{0.15})$      &  $ +1.85\ (\textcolor{magenta}{0.18})$     \\
%       \bottomrule
%     \end{tabular} 
%         \label{component}
%                   \vspace{-0.1in}
%   \end{table}

\section{Conclusion}

In this paper, we take the first attempt to study the self-supervised graph representation learning in the mixed-curvature Riemannian space, and present a novel \textsc{SelfMGNN}.
Specifically, we first construct the mixed-curvature space via Cartesian product  of Riemannian manifolds
and design hierarchical attention mechanisms within and among component spaces to learn graph representations in the mixed-curvature space.
%In this mixed-curvature space, 
%we reveal different Riemannian views by the designed Riemannian projector and propose the dual contrastive approach, i.e., 
Then, we introduce the
single-view
and cross-view contrastive learning to learn graph representations without labels.
Extensive experiments show the superiority of \textsc{SelfMGNN}.
% outperforms state-of-the-art baselines. % on several benchmark datasets.

 % \begin{table}
 %  \footnotesize
 %    \centering
 %          \caption{The summary of related work}
 %          \vspace{-0.1in}
 %    \begin{tabular}{c| c|c}
 %      \toprule
      
 %      \bottomrule
 %    \end{tabular} 
 %  \end{table}

%% file: appendix.tex
%!TEX root = ./SelfMGNN.tex

In the technical appendix, we provide further details for the proof and derivations as well as the experiments.

\section{A. On the Attentional Aggregation}
In this section, we first start with attentional aggregation in the Euclidean space, 
and then elaborate on the necessary operations in the $\kappa$-stereographic model $\mathcal M$ and generalize the attentional aggregation in the Euclidean space to the $\kappa$-stereographic model.
Finally, we prove the \emph{Theorem 1 ($\kappa$-left-matrix-multiplication as aggregation)}.

\subsection{Attentional aggregation in the Euclidean space}
In this part, we show that attentional aggregation is a linear combination of the  feature vectors with the learnable weights and left-matrix-multiplication essentially performs the attentional aggregation.

%The attentional aggregation is the summation of feature vectors with the learnable weights. 
Given a target node $i$ and its neighbor node set $\mathcal N_i$, 
the encoding of the target $\boldsymbol z_i$  is updated as 
$\sum\nolimits_j A_{ij}\boldsymbol z_j$,
where $j \in \Omega$ and $\Omega =\mathcal N_i \cup i$. We add a self-loop in particular to keep the information of the node itself, and $A_{ij}$ denotes the attentional weight to be learned \cite{WuPCLZY21,velickovic2018graph}.
Obviously, the attentional aggregation can be rewritten as a linear combination  $\boldsymbol L(\cdot, \cdot)$. 
Specifically, the linear combination is defined as $\boldsymbol L(c(A_{ij}), \boldsymbol z_j)=\sum\nolimits_j c(A_{ij})\boldsymbol z_j$,
where $c(A_{ij})$ is the coefficient function to output a real coefficient scaling the corresponding $\boldsymbol z_j$. We have $c(A_{ij})=A_{ij}$ for the attentional aggregation above.

Let us consider left-matrix-multiplication of $\mathbf Z\in \mathbb R^{N \times D}$ by $\mathbf A\in \mathbb R^{N \times N}$, i.e., we have $\mathbf A\mathbf Z \in \mathbb R^{N \times D}$.  
The $i^{th}$ row of $(\mathbf A\mathbf Z)$ is given as follows:
\begin{equation}
A_{i1}\boldsymbol z_1+A_{i2}\boldsymbol z_2 + \cdots + A_{ij}\boldsymbol z_j+ \cdots +A_{iN}\boldsymbol z_N,
\end{equation}
where $ \boldsymbol z_j$ is the $j^{th}$ row of $\mathbf Z$.
In fact, the row-wise left-matrix-multiplication is given by the linear combination $(\mathbf A\mathbf Z)_{i \bullet}=\boldsymbol L(c(A_{ij}), \boldsymbol z_j)$ with $\mathbf A$, performing the attentional aggregation.
Thus, we can give the lemma as follows:
\newtheorem*{lemma}{Lemma 1 (Left-matrix-multiplication as attentional aggregation)}
\begin{lemma}
Given $\mathbf{Z}\in \mathbb R^{N \times D}$ holding encoding vectors in its row and weights $\mathbf{A}\in \mathbb R^{N \times N}$, 
the left-matrix-multiplication of $\mathbf{Z}$ by $\mathbf{A}$ updates encoding vectors in $\mathbf{Z}$ with attentional aggregation.
\end{lemma}

\subsection{Operations in the $\boldsymbol \kappa$-stereographic model}
In this part, we first review the notion of \emph{midpoint} in the $\kappa$-stereographic model, and then generalize the Euclidean left-matrix-multiplication  to the \emph{$\kappa$-left-matrix-multiplication} in the $ \kappa$-stereographic model with the \emph{midpoint}.

The midpoint in the Euclidean space is intuitive, however, it is nontrivial in the $\kappa$-stereographic model as the manifold is curved.
We give the definition of the (weighted) midpoint in the $\boldsymbol \kappa$-stereographic model as follows:
\newtheorem*{def2}{Definition 1 (Midpoint in the $\boldsymbol \kappa$-stereographic model)} 
\begin{def2}
Given a set of  $\kappa$-stereographic vectors $\{\mathbf{x}_{i}\}_{i=1}^n$, and weights $\boldsymbol{\alpha} \in \mathbb R^n$,
the weighted midpoint in the $\kappa$-stereographic model is calculated via $\mathbf{mid}_{\kappa}\left( \{\mathbf{x}_{i}\}_{i=1}^n ; \boldsymbol{\alpha}\right)$ as follows:
\begin{equation}
%\resizebox{0.91\hsize}{!}{$
\mathbf{mid}_{\kappa}\left( \{\mathbf{x}_{i}\}_{i=1}^n ; \boldsymbol{\alpha}\right)=\frac{1}{2} \otimes_{\kappa} \left(\sum_{i=1}^{n} \frac{\alpha_{i} \lambda_{\mathbf{x}_{i}}^{\kappa}}{\sum_{j=1}^{n} \alpha_{j} (\lambda_{\mathbf{x}_{j}}^{\kappa}-1)} \mathbf{x}_{i}\right),
%$}
\end{equation}
where  $\lambda_{\mathbf{x}_{j}}^{\kappa}=4\left(1+\kappa||\mathbf x||_2^2\right)^{-2}$ is the conformal factor.
\end{def2}
\noindent Note that, the midpoint in the $\kappa$-stereographic model is essentially a linear combination regulated with a $\kappa-$scaling.

With the geometry of the $\kappa$-stereographic model, we give the definition  of $\kappa$-left-matrix-multiplication following \citet{BachmannBG20}, the generalization of left-matrix-multiplication in the Euclidean space, below:
\newtheorem*{def0}{Definition 2 ($\boldsymbol \kappa$-left-matrix-multiplication)} 
\begin{def0}
Given $\mathbf{Z}\in \mathbb R^{N \times D}$ holding $\kappa$-stereographic vectors in its row and weights $\mathbf{A}\in \mathbb R^{N \times N}$, 
the $\kappa$-left-matrix-multiplication of $\mathbf{Z}$ by $\mathbf A$ is defined as follows:
\begin{equation}
\left(\mathbf{A} \boxtimes_{\kappa} \mathbf{Z}\right)_{i \bullet}:= A \otimes_{\kappa} \mathbf{mid}_{\kappa}\left( \{ \mathbf{Z}_{i \bullet}\}_{i=1}^n ; \mathbf{A}_{i \bullet}\right),
\end{equation}
where $A=\sum_{j} \mathbf A_{i j}$, $\mathbf{mid}_{\kappa}$ denotes midpoint in the $\kappa$-stereographic model.
\end{def0}

\subsection{Attentional Aggregation in  $\boldsymbol \kappa$-stereographic Model}
In this part, we show that the $\kappa$-left-matrix-multiplication performs the attentional aggregation in $\kappa$-stereographic model.
In other words, we prove \textbf{Theorem 1} in the subsection of \emph{attentional aggregation layer} in our paper.

With the formal definition of the linear combination, we rewrite the  \textbf{Theorem 1} equivalently as follows:
\newtheorem*{thm1}{Theorem 1 ($\boldsymbol \kappa$-left-matrix-multiplication as attentional aggregation)}
\begin{thm1}
Let rows of $\mathbf{H}$ hold the encoding $\boldsymbol z_{\mathcal M_i}$, (linear transformed by $\mathbf{W}$)
and $\mathbf{A}$ hold the attentional weights,
the  $\kappa$-left-matrix-multiplication $\mathbf{A} \boxtimes_{\kappa} \mathbf{H}$ performs the attentional aggregation over the rows of $\mathbf{H}$, i.e., 
$\mathbf{A} \boxtimes_{\kappa} \mathbf{H}$ is the row-wise linear combination of  $\mathbf{H}$ with respect to attentional weight $\mathbf{A}_{ij}$: 
\begin{equation}
(\mathbf{A} \boxtimes_{\kappa} \mathbf{H})_{i \bullet}=\boldsymbol L(c_{\text{stereo}}(\mathbf{A}_{ij}), \mathbf{h}_{j}),
\end{equation}
% \begin{equation}
% (\mathbf{A} \boxtimes^{\kappa} \mathbf{H})_{i \bullet}=\oplus^{\kappa}_{j\in \Psi}(\mathbf{A}_{ij} \otimes^{\kappa} \mathbf{H}_{i \bullet}),
% \end{equation}
where  $\mathbf{h}_{j}$ is the $j^{th}$ row of $\mathbf H$, $j$ enumerates the index set $\Psi$, $\Psi=i \cup \mathcal N_i$ and $\mathcal N_i$ is the neighbors of $i$ on the graph. $c_{\text{stereo}}(\cdot)$ is the function to output the coefficient in the $\kappa$-stereographic model.
\end{thm1}
\begin{proof}
Recall the design of the (intra-component) attentions in our paper:
$\mathbf A$ is given as $\hat{\mathbf A}+ \mathbf I$,
where $\hat{\mathbf A}$ is filled with softmax values in its row, and $\mathbf I$ is the identity matrix to keep the initial information of the node itself.
That is, we have the row sum, $A=2$.
Then, with the definitions of \emph{$\kappa$-left-matrix-multiplication} and \emph{midpoint in the $\kappa$-stereographic model}, we give the derivation as follows:
\begin{equation}
\begin{aligned}
   &(\mathbf{A} \boxtimes_{\kappa} \mathbf{H})_{i \bullet}\\
=&A \otimes_{\kappa} \mathbf{mid}_{\kappa}\left( \{ \mathbf{h}_j\}_{j=1}^n ; \{\mathbf A_{ij}\}_{j=1}^n\right)\\
=&2\otimes_{\kappa} \frac{1}{2} \otimes_{\kappa} \left(\sum_{j=1}^{n} \frac{\mathbf A_{ij} \lambda_{\mathbf{h}_{j}}^{\kappa}}{\sum_{l=1}^{n} \mathbf A_{il} (\lambda_{\mathbf{h}_{l}}^{\kappa}-1)} \mathbf{h}_{j}\right)\\
=&\sum_{j=1}^{n} \frac{\mathbf A_{ij} \lambda_{\mathbf{h}_{j}}^{\kappa}}{\sum_{l=1}^{n} \mathbf A_{il} (\lambda_{\mathbf{h}_{l}}^{\kappa}-1)} \mathbf{h}_{j}\\
=&\boldsymbol L(c_{\text{stereo}}(j), \mathbf{h}_{j}),\\
\end{aligned}
\end{equation}
where the coefficient function is given as $c_{\text{stereo}}(j)=\frac{1}{C}\mathbf A_{ij} \lambda_{\mathbf{h}_{j}}^{\kappa}$, and $C=\sum_{l=1}^{n} \mathbf A_{il} (\lambda_{\mathbf{h}_{l}}^{\kappa}-1)$.
As shown above, with the well-designed attention mechanism, 
we eliminate the $\kappa$-scaling and make the $\kappa$-left-matrix-multiplication to be the linear combination with respect to the learnable attentions, 
performing the attentional aggregation.
\end{proof}

 \begin{table}
    \centering
    \setcounter{table}{1}
          \caption{The statistics of  the datasets.}
    \begin{tabular}{ p{1.5cm}<{\centering} |   p{2cm}<{\centering}  p{2cm}<{\centering}   p{1.2cm}<{\centering}}
      \toprule
\textbf{Dataset}&   \#(Node)   &  \#(Links) &   \#(Labels)      \\
\toprule
\textbf{Citeseer}& $\ \ 3,327$ & $\ \ \ \ 4,732$ & $6$ \\
\textbf{Cora}      & $\ \ 2,708$ & $\ \ \ \ 5,429$ & $7$ \\
\textbf{Pubmed}& $19,717$ & $\ \ 44,338$ & $3$ \\
\textbf{Amazon}& $13,381$ & $245,778$ & $10$ \\
\textbf{Airport}  & $\ \ 1,190$ & $\ \ 13,599$ & $4$ \\
      \bottomrule
    \end{tabular} 
        \label{statistics}
  \end{table}

\section{B. Experimental Details}
In this section, we give further experimental details, including data \& code  and implementation notes, in order to enhance the \emph{reproducibility}.
\subsection{Data and Code}
\subsubsection{Data} The datasets used in this paper are publicly available, i.e., Citeseer, Cora, Pubmed, Amazon and Airport. We briefly describe these datasets as follows:
\begin{itemize}
   \item Citeseer , Cora and Pubmed are the widely used citation networks, where nodes represent papers, and edges represent citations between them.
   \item The Amazon is a co-purchase graph, where nodes represent goods and edges indicate that two goods are frequently bought together. 
   \item The Airport is an air-traffic graph,  where nodes represent airports and edges indicate the traffic connection between them.
\end{itemize}
We list the statistics of  the datasets in Table \ref{statistics}.

\subsubsection{Code} We submit the source code of an instance implementation of \textsc{SelfMGNN} in a ZIP named \emph{Code}, and will publish the source code after acceptance.  
% \subsection{Running Environment}
% All experiments were conducted on the a CentOS server with two $11$G Nvidia RTX 2080Ti and $128$G RAM.
\subsection{Implementation Notes}
In \textsc{SelfMGNN}, we stack the attentive aggregation layer twice to learn the  component embedding. 
We employ a two-layer MLP$_\kappa$ in the Riemannian projector to reveal the Riemannian views for the self-supervised learning. 
In the experiments, we set the weight $\gamma$ to be $1$, i.e., the single-view and cross-view contrastive learning are considered to have the same importance.
The grid search is performed over the learning rate in $[0.001, 0.003, 0.005, 0.008, 0.01]$ as well as the dropout probability in $[0, 0.8]$ with the step size of $0.1$.

For all the comparison model, we perform a hyper-parameter search on a validation set to obtain the best results, 
and the $\kappa$-GCN is trained with positive curvature in particular to evaluate the representation ability of the spherical space.
We set the dimensionality to be $24$ for all the models for the fair comparison.
Note that, in \textsc{SelfMGNN}, the component space can be set to arbitrary dimensionality, whose curvature and importance are learned from the data, and thereby we  construct a mixed-curvature space of  any dimensionality, matching the curvatures of any datasets.
%\subsection{Parameter Settings}

% Use \bibliography{yourbibfile} instead or the References section will not appear in your paper